\documentclass{article} 
\usepackage{iclr2021_conference,times}

\usepackage{amsmath,amsfonts,bm}

\def\eqref#1{equation~\ref{#1}}

\def\1{\bm{1}}

\DeclareMathAlphabet{\mathsfit}{\encodingdefault}{\sfdefault}{m}{sl}
\SetMathAlphabet{\mathsfit}{bold}{\encodingdefault}{\sfdefault}{bx}{n}

\usepackage[utf8]{inputenc}
\usepackage[hidelinks]{hyperref}
\usepackage{url}
\usepackage{booktabs}

\title{NeuralLog: a Neural Logic Language}

\author{Victor Guimarães \And Vítor Santos Costa}

\author{Victor Guimarães \\
CRACS and DCC/FCUP \\
Universidade do Porto \\
Porto, Portugal \\
\texttt{victorguimaraes13@gmail.com} \\
\And
Vítor Santos Costa \\
CRACS and DCC/FCUP \\
Universidade do Porto \\
Porto, Portugal \\
\texttt{vsc@dcc.fc.up.pt}
}

\usepackage{subcaption}
\usepackage{multirow}
\usepackage{amsmath}

\usepackage[ruled]{algorithm}
\usepackage{algpseudocode}
\usepackage{calc}
\usepackage{amssymb}

\algrenewcommand\algorithmicrequire{\makebox[\widthof{\textbf{Output:}}][l]{\textbf{Input:}}}
\algrenewcommand\algorithmicensure{\makebox[\widthof{\textbf{Output:}}][l]{\textbf{Output:}}}
\algrenewcommand\textproc{}

\usepackage{tikz}
\usepackage{tikzscale}
\usetikzlibrary{arrows,automata}
\usetikzlibrary{shapes,decorations.pathmorphing}
\usetikzlibrary{positioning}
\usetikzlibrary{calc}

\newcommand{\norm}[1]{\left\lVert#1\right\rVert}
\newtheorem{example}{Example}

\iclrfinalcopy 
\begin{document}

\maketitle

\begin{abstract}
   Application domains that require considering relationships among objects which have real-valued attributes are becoming even more important.
   In this paper we propose NeuralLog, a first-order logic language that is compiled to a neural network. The main goal of NeuralLog is to bridge logic programming and deep learning, allowing advances in both fields to be combined in order to obtain better machine learning models.
   The main advantages of NeuralLog are: to allow neural networks to be defined as logic programs; and to be able to handle numeric attributes and functions.
   We compared NeuralLog with two distinct systems that use first-order logic to build neural networks. We have also shown that NeuralLog can learn link prediction and classification tasks, using the same theory as the compared systems, achieving better results for the area under the ROC curve in four datasets: Cora and UWCSE for link prediction; and Yelp and PAKDD15 for classification; and comparable results for link prediction in the WordNet dataset.
\end{abstract}

\section{Introduction}

Deep learning has been remarkably successful on a wide range of tasks \citep{yLeCun98}. However, most of those task are based on propositional numeric inputs, assuming that there are no relation between the examples.
Logic programming \citep{rBrachman04}, on the other hand, uses logic programs to describe and to reason about structured and multi-relational data, but it struggles to deal with uncertainty and noise, two factors inherent to real world problems.

In order to overcome these limitations, we propose \emph{NeuralLog}, a first-order logic language that is compiled to a neural network. The main goal of NeuralLog is to bridge logic programming and deep learning in order to exploit the advantages of these fields in both discrete and continuous data.

Another main contribution of NeuralLog is to allow one to abstract the design of deep neural networks by the use of a logic language, which would facilitate the use of deep learning models by people that are not familiarized with common programming languages, since logic languages have been used for a long time as modelling language for different domains.

In addition, NeuralLog allows to define neural networks to address relational tasks. It also supports numeric attributes and the use of numeric functions, thus providing the ability to create generic neural networks;
in contrast with most of the works in this field, that are restricted at combining logic with neural networks on discrete domains composed by relations among entities.
Furthermore, the logic program is mapped to a neural network in such a way that the logic of the program can be easily identified in the structure of the network and vice versa, making the model easier to interpret.

We compare NeuralLog with two state-of-the-art systems in two very different domains: link prediction with TensorLog \citep{wCohen17} and classification with RelNN \citep{sKazemi18s}. Our results show that NeuralLog performs as well as, and often better than, these two other systems on tasks they were designed for, and support our claim that NeuralLog is a significant contribution towards a flexible neural relational system.

The remaining of the paper is organised as follows: in Section \ref{sec:related_work}, we show some related works; in Section \ref{sec:background}, we give the background knowledge to follow this work; in Section \ref{sec:neurallog} we present NeuralLog; and the performed experiments in Section \ref{sec:experiments}; finally, we present our conclusion and directions of future work in Section \ref{sec:conclusion}.

\section{Related Work} \label{sec:related_work}

There are many ways to combine first-order logic with neural networks, from using logic features as input to other machine learning models \citep{hLodhi13,dTirtharaj18} to the use of the logic rules to define the structure of the neural network \citep{mFranca14,aFadja17,gSourek18,hDong18,kNavdeep19}.

TensorLog \citep{wCohen17} is a system that performs inference of first-order logic by using numeric operations. It represents the logic facts as matrices and performs logic inference through mathematical operations on those matrices. Another well-known work that relies on matrix representations is Logic Tensor Network \citep{lSerafini16}, which uses a tensor network \citep{rSocher13} to predict the confidence of the logic formulas.

NeuralLog uses the same underlying idea of representing logic facts as matrices and mathematical operations to compute logic rules.
However, it differs from TensorLog in two main aspects: (1) in the way how the logic rules are compiled to a neural network; and (2) by treating numeric attributes as logic terms that can be flexibly manipulated and also given as input to numeric functions. This allows NeuralLog to be more flexible than TensorLog. 

Similar to TensorLog, RelNN uses matrices and vectors operations to perform relational learning, using convolution neural networks \citep{sKazemi18s}. They use the number of logic proofs of rules as features, that can also be combined into other rules, multiplied by weights, added to bias, and passed as input to activation functions.
Both TensorLog and RelNN can tune the weights of facts to better predict a set of examples.



We compared NeuralLog with TensorLog and RelNN, in particular, because in those systems the structure of the neural network clearly reflects the logic theory.

It is important to notice that the use of first-order logic is not the only way to allow neural networks to be applied to relational tasks. However, the other approaches, although promising, are beyond the scope of this paper.

\section{Background Knowledge} \label{sec:background}

In this work we use first-order logic to represent and learn neural network models in multi-relational tasks. The knowledge is represented as a set of facts that is called knowledge base (KB).
A fact is composed by a \emph{predicate} (relation) and, possibly, a set of constants (entities). A fact can be represented as $p(s_1, \dots, s_n).$, where $p$ is the predicate, and $s_i$ are the constants; and states that there is a relation of type $p$ among the constants $s_i$. A fact may also have no constants, in this case, we say the fact is propositional; for instance $rain$ states that it rains.

To reason about this knowledge we use Horn Clauses \citep{aHorn51}. A Horn Clause, also referred as a rule, has the form:
	$p_0(.) \leftarrow p_1(.) \wedge \dots \wedge p_n(.).$
Where $p_i$ are predicates and $(.)$ are lists of \emph{terms}, a term can be either a \emph{constant} or a \emph{variable}. A constant is represented by a string starting with a lower case letter, while variables start with an upper case letter. The predicate followed by its terms, in the clause, is called a literal. A literal is an atom or the negation of an atom. Although we do not consider negation in this work, we will use the term literal to refer to atoms in the body of a rule.
The atom in the left-hand side of the rule is called the head, and the conjunction of literals in the right-hand side is called the body.
The arity of a predicate is the number of terms a literal of this predicate accepts. A predicate $p$ of arity $n$ is also represented as $p/n$. In this work, we do not consider first-order logic functions, only numeric functions, as will be described in more detail in the next section. We also limit the facts to have arity of at most two.

An atom is proved (considered to be true) whenever it matches a fact in KB or it can be proved by a rule. A variable can be replaced by a constant in order to make the atom to match a fact. A rule proves the atom in its head whenever there is a substitution of its variables that makes \textbf{all} proves all the literals in its body, possible by using other rules. A negated literal ($not \ \mathbf{A}$) is considered to be true whenever we cannot prove $\mathbf{A}$.
We base the NeuralLog syntax on DataLog \citep{sAbiteboul95}, adding the weighted fact syntax from ProbLog \citep{lDeRaedt07}.

Generically, an artificial neural network is a directed graph of layers, where each layer receives as input the results of its income layers and computes its output based on its inputs and its internal parameters.
The learning process consists of finding set of the parameters of the layers that better describes the training data.
This is usually done by a gradient descent algorithm \citep{sHaykin94}.


\section{NeuralLog} \label{sec:neurallog}

In this section, we present NeuralLog, a first-order logic language that describes neural networks to perform logic inference.

Since neural networks process numeric inputs, it is convenient to describe the KB in a numeric form. 
More precisely, we are given a logic program that can be seen as a triple $(E, F, R)$, where $E$ is the set of entities available in the domain; $F$ is the set of \emph{facts}; and $R$ is the set of rules. Each fact may have at most two terms and always has a weight $w$ (by default 1), that might be omitted.

Similarly to \citep{wCohen17}, NeuralLog stores knowledge as (sparse) matrices, to be evaluated through algebraic operations on those matrices. First, we construct an index for $E$ by assigning to each entity an integer value in $[0, n)$, where $n = |E|$. Next, for each binary predicate $p$, we create a matrix $P \in {\rm I\!R}^{n \times n}$ where $P_{ij} = w$ if there is a fact $p(e_i, e_j)$ with weight $w$ in $F$, where $i$ and $j$ are the indices of the entities $e_i$ and $e_j$, respectively; otherwise, $P_{ij} = 0$. Predicates of arity $1$ (unary) are mapped to vectors; and propositional predicates to scalars. An entity (or constant) is represented by an one-hot vector, having the value of $1$ for the entry correspondent to the index of the entity and the value of $0$ for every other entries.

\paragraph{Real-Valued Data} One of our goals in the design of NeuralLog was to support numeric value data. Naively introducing real values as any other constant in $E$ could lead to very large matrices. 
Instead, we assume that real values are bounded to entities in KB as attributes of the entities. Our implementation uses two vectors ($\mathbf{p_v}, \mathbf{p_w} \in {\rm I\!R}^n$) for each attribute predicate $p$. Where $\mathbf{{p_v}}_i = a$ and $\mathbf{{p_w}}_i = w$ if there is a fact $p(e_i, a)$ with weight $w$ in $F$, and $i$ is the index of the entity $e_i$; otherwise, $\mathbf{{p_v}}_i = \mathbf{{p_w}}_i = 0$. As an example, the predicate $length$ will consist of facts such as $length(x, 1.5)$, informing that entity $x$ has an attribute \emph{length} of value $1.5$.

\subsection{Inference}

The key idea of NeuralLog is to transform a logic program into a neural network such that both rule inference and the integration of different
rules can be performed through matrix operations.

The goal of NeuralLog is to find all possible entities that might replace the variable $Y$, associated with the entity $X = \mathbf{a}$, with relation $p$, for a given a query of the form $?- p(X, Y)$; with respect to the knowledge base and clauses.

To briefly explain the translation, imagine a KB composed of facts for a single binary predicate $p/2$. The implementation is a matrix $P$, with an one-hot row vector $\mathbf{a}$ representing a given entity in KB, as described above. Given $X = \mathbf{a}$ and a query $?- p(X, Y)$, $Y = \mathbf{a} P$ describes the values of $Y$. In the other direction, if we would like to compute the vector for the entities $X$ related to $a$ where $Y = \mathbf{a}$, we would do so by transposing the matrix $P$ such as $X = a P^\intercal = (Pa^\intercal)^\intercal$. We extend this method to unary and attribute facts by using the element-wise multiplication. 

Finally, the result of the query $?- p(X, Y)$, for $X = \mathbf{a}$ will be the set of facts, for each possible entities in KB, where the weight of each fact $p(a, e_i)$, where $e_i$ is the entity assigned to position $i$, would be $Y_i$, from the resulting vector $Y$.

\subsection{Rule Graph Construction}

In this section, we extend the previous inference approach in order to compute the output value of arbitrary rules. In order to do so, we start by creating an undirected graph from the rule, where the nodes are the terms in the rule; and there is an edge between two nodes if they both appear together as terms in a literal of the body of the rule. Then, we define the last term in the head of the rule as the output (destination) node and the remaining terms in the head as inputs (sources) nodes.

Assuming that edges represent influence, we perform a breadth-first search in this graph, in order to find all the paths from the source node to the destination node. We constraint ourself with rules whose arity of the head are greater than zero and we use a topological order to disallow back-edges and loops.

\begin{figure}
	\center
	\begin{subfigure}{0.45\textwidth}
		\center
		\resizebox{\textwidth}{!}{
			\begin{tikzpicture}[->,>=stealth',shorten >=0pt, auto, node distance=1.5cm, semithick, minimum size=10mm]
				\tikzstyle{every state}=[fill=none, draw=black, text=black]
				\tikzstyle{every node}=[circle, fill=none, draw=black, text=black, align=center]

				\node[draw] (X)	{$X$};
				\node[draw] (Z)	[right = of X] {$Z$};
				\node[draw] (Y)	[right = of Z] {$Y$};
				\node[draw=none] (w)	[right = of Y,xshift=-1.0cm] {$w$};

				\node[draw] (V)	[above = of Z] {$V$};
				\node[draw] (U)	[below = of Z] {$U$};

				\path
				(X) edge [bend left=30] node[draw=none,yshift=-0.5cm] {$p_0(X, Z)$} (Z)
				(X) edge [bend right=30] node[draw=none,yshift=-1.3cm] {$p_1(X, Z)$} (Z)

				(Z) edge node[draw=none,yshift=0cm] {$p_2(Z, Y)$} (Y)

				(X) edge [bend left=30] node[draw=none,xshift=-0.0cm] {$p_3(X, V)$} (V)
				(U) edge [bend right=30] node[draw=none,xshift=1.25cm,yshift=-1.4cm] {$p_4(U, Y)$} (Y)

				(V) edge [bend left=30, dashed] node[draw=none,xshift=-0.0cm] {$any(V, Y)$} (Y)
				(X) edge [bend right=30, dashed] node[draw=none,xshift=-1.25cm,yshift=-1.5cm] {$any(X, U)$} (U)
				(Z) edge [loop below] node[draw=none,yshift=0.4cm] {$r5(Z)$} (Z)
				;
			\end{tikzpicture}
		}
		\caption{The DAG representation of the rule} \label{fig:rule_graph}
	\end{subfigure}
	\hfill
	\begin{subfigure}{0.45\textwidth}
		\center
		\resizebox{\textwidth}{!}{
			\begin{tikzpicture}[-,>=stealth',shorten >=0pt, auto, node distance=1.5cm, semithick, minimum size=10mm]
				\tikzstyle{every state}=[fill=none, draw=black, text=black]
				\tikzstyle{every node}=[rectangle, fill=none, draw=black, text=black, align=center,draw=none]

				\node (Z)	{$Z$};
				\node (Z')	[right = of Z,xshift=-0.0cm] {$Z$};
				\node (V)	[right = of Z',xshift=-0.0cm] {$V$};
				\node (U)	[right = of V,xshift=-0.0cm] {$U$};

				\node (X) at ($(Z')!0.5!(V) + (0, 2.5cm)$)	{$X$};

				\node (Z'')		[below = of Z,yshift=0.45cm] {$Z$};
				\node (Z''')	[below = of Z',yshift=0.45cm] {$Z$};

				\node (Y)		[below = of Z'',yshift=0.45cm] {$Y$};
				\node (Y')		[below = of Z''',yshift=0.45cm] {$Y$};
				\node (Y'')		[below = of V,yshift=0.45cm] {$Y$};
				\node (Y''')	[below = of U,yshift=0.45cm] {$Y$};

				\path
				(X.225) edge node[yshift=0.0cm,xshift=-1.0cm] {$p_0$} (Z.90)
				(X.-112.5) edge node[yshift=0.0cm,xshift=-0.5cm] {$p_1$} (Z'.90)
				(X.-67.5) edge node[yshift=-1.0cm,xshift=0.2cm] {$p_3$} (V.90)
				(X.-45) edge node[yshift=-1.0cm,xshift=1.2cm] {$any$} (U.90)

				(Z) edge node[xshift=-0.2cm] {$p_5$} (Z'')
				(Z') edge node[xshift=-0.2cm] {$p_5$} (Z''')
				(V) edge node[xshift=-0.05cm] {$any$} (Y'')
				(U) edge node[xshift=-0.2cm] {$p_4$} (Y''')

				(Z'') edge node[xshift=-0.2cm] {$p_2$} (Y)
				(Z''') edge node[xshift=-0.2cm] {$p_2$} (Y')
				;
			\end{tikzpicture}
		}
		\caption{All paths from X to Y represented as a tree} \label{fig:path_tree}
	\end{subfigure}
	\caption{Example of the DAG representation (on the left-hand side) and the found paths (on right-hand side) of the rule} \label{fig:rule_example}
\end{figure}
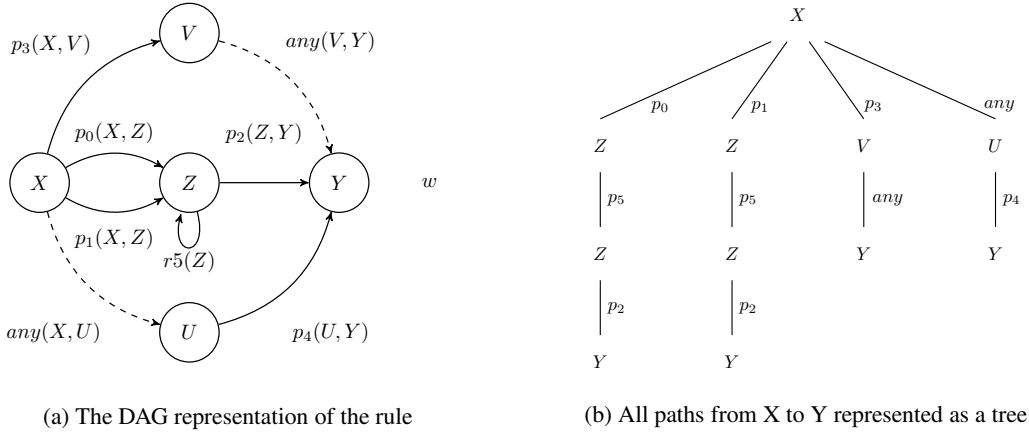

Figure \ref{fig:rule_example} shows an example of the DAG and the found paths (as a tree) of the rule below:
\begin{equation*}
	target(X, Y) \leftarrow p_0(X, Z) \wedge p_1(X, Z) \wedge p_2(Z, Y) \wedge p_3(X, V) \wedge p_4(U, Y) \wedge p_5(Z) \wedge w.
\end{equation*}

In this case, our neural network (NN) will be a function $Y = f(X)$. A first step shows edges $X-Z$ and $X-V$. In a second step we obtain $Z-Y$; as we do not allow cycles, $V$ has no path to the output, and $U$ is disconnected from the network.

In order to improve the compiled network, by accounting by the influence of disconnected terms, we use a special $any/2$ predicate, inspired by TensorLog \citep{wCohen17}, that is essentially true for any pair of terms. There are two possible cases where we use this predicate:
\begin{enumerate}
	\item connected nodes that do not have a path to target can still influence the target ($Y$): this is implemented by adding an edge from the node to the output, as shown in $V-Y$;
	\item unconnected nodes may also influence the output; we use $any/2$ to connect the input to the node, as shown in $X-U$, in order to consider the relation from $U$.
\end{enumerate}
Notice that the second case introduces new partial paths and we need to restart the algorithm to propagate these paths. The process repeats until no new edges are needed to be added. The algorithm always terminates as the number of steps is bounded by the number of nodes. The algorithm to find those paths are described in Appendix \ref{sec:rule_algorithms}.

Given the found paths, we construct a directed acyclic graph (DAG) representation of the rule. Equal edges in different paths are collapsed into a single edge; and different incoming $any/2$ edges to the destination are represented as a single $any/n$ literal, where $n$ is the number of distinct terms appearing in all these edges and the destination is the last term. All unary predicates (which would represent a loop in the graph) is then added to their correspondent edges.

The inference of the rule, given its inputs, is represented by the result of the destination node in the DAG, multiplied by any disconnected grounded literal in the body of the rule. Those are grounded literals that do not appear in any path, which include the propositional literals. Since they do not have variables, they are represented as scalars. It is important to notice that, in the example, $w$ is a propositional predicate, represented as a scalar, and not a term.

In order to compute the result of a node we combine the values of its incoming edges by an element-wise multiplication (representing a logic \textbf{AND}); then we combine the values of the unary predicates of the node, if any, in the same order they appear in the rule. 

In order to compute the value of an edge (which represents a literal) we multiply the vector representing the input term of the edge by the matrix representation of the edge's predicate. 
If the predicate is represented as a vector (in the case of unary predicate), we perform an element-wise multiplication, instead. If it is an attribute predicate, we perform the element-wise multiplication between the input term, the weight and the attribute vectors.
Since we need the values of the incoming nodes to compute the value of the current node, we recursively transverse the DAG until we reach a source node, whose value in given from the input. 

The $any/n$ predicate has a special treatment. First, it is important to notice that the $any/n$ predicate can only appear in two specific situations, as enumerated above: (1) connecting the end of a path to the destination; or (2) connecting an input term to the beginning of a path that will lead to the destination (which was computed from the destination to the input and then reversed).	

In order to compute the first case of the $any/n$ predicate, for each term of the predicate, except the last one, we compute the vector resulting from the term and sum it to get a scalar. For the last term (which is always the destination term of the rule), we compute the vector resulting from this term, based on its unary predicates, by passing a $1$ vector as input value for the predicates; then, we multiply it by the scalar representation of the previous terms.

Intuitively, the result represents the multiplication of the sum of the results of each (any) term by the results of each (any) entities of the last term. The final result is combined with other possible edges arriving at the destination term. In this case, the unary predicate is not combined with the final result, since it has been already combined by the $any/n$ predicate.

The second case of the $any/n$ occurs when we have edges connected to the destination that do not lead to any input. As such, the first $n - 1$ terms are input terms and the last term is the beginning of the path that leads to the destination. In this case, we pass a $1$ vector, representing the input of the term. Then, we apply the existent unary predicates, if any, to it. In numeric terms, the $1$ vector represents the union of all (any) logic constants. In logic terms, the node represents a free variable in the rule. Thus, the result of the rule does not depend on it, meaning that it is valid for any constant in this variable.

In addition, NeuralLog allows the use of numeric functions. The functions are applied, in order of appearance in the rule, to the computed vector of its inputs.
There is no explicit differentiation between functions and predicates in NeuralLog language. The difference is that predicates have facts defined in the KB and the computation is the multiplication of the input terms with the matrix (or vector) representation of the facts of the predicates, while the result of function predicates is the application of the function to the input terms.
These functions can be any differentiable function, including other network layers or even whole neural network models.

If a relation involves a constant, we multiply the input (output) variable with the vector representation of the input (output) constant. It  represents a logic \textbf{AND} of the input (output) variable with the input (output) constant. We do the same for the $any/n$ predicate.

\subsection{Network Construction}

We have explained how we represent the knowledge data and how to compute an individual rule represented by a Horn clause. In this section, we show how to construct a neural network from a first-order logic program (a set of logic rules and facts).

In order to build the network we start by creating a \emph{Literal Layer} for each target predicate. This layer receives as input all the rules and facts that have the target predicate in their heads.
The output of a literal layer is the sum of the results of the \emph{Fact Layer}, representing the matrix of facts in the KB and the \emph{Rule Layers} of the target literal, similar to \citep{gSourek18}; given the input terms of the literal. The sum of the results represents a logic \textbf{OR}.

The output of the \emph{Fact Layer} is the input vector multiplied by the matrix (or vector) of the weights of the facts in the KB. For attribute facts, their weight and value vectors are combined by an element-wise multiplication.

Recursively, for each rule connected to the target predicate node, we create a \emph{Rule Layer}, and connect, as its inputs, the corresponding \emph{Literal Layers} from the rule's body. We represent function predicates in the body of the rule as \emph{Function Layers}, the output of a function layer is the application of the function to the vector representation of its inputs terms. The computation of the rule layer is performed as explained in the subsection above, using the computation performed by the literal layers, instead of multiplying the matrices of the facts. We repeat this process until no more layers are created.

In this way, we unfold the network from the structure of the logic rules, similar to \citep{aFadja17}. However, our language allows the use of attribute predicates and functions. Whenever a recursion is found in the rules, it is unfolded until a pre-defined depth.
Since the rule inference operation is differentiable, the unfolding of the rules, and thus the whole network, is differentiable.

The \emph{Literal Layers} of the target predicates receive as input the vector corresponding to their input terms, and represent the inputs of the neural network; while the result of these layers are the output of the network. Note that if we want to compute an inverted literal, by going from the last term of the literal to the first, we can simply use the transpose matrix of the relation, for the fact layer; and compute the reversed paths, for the rule layers.
The depth of the neural network is defined by the logic program, and a hierarchy of abstract concepts can be created by the use of predication invention.

In order to learn a task from data, we allow the user to specify predicates whose weights will be adjusted from examples. In this way, the weights of those predicates will become parameters to be learned. After training, we can extract the learned weights to update the logic program that generated the network. We show some examples and the algorithms to find the paths in the rules in appendix.

NeuralLog is implemented in Python \citep{vRossum09} using TensorFlow \citep{mAbadi15} as backend and its source code is public available\footnote{\url{https://github.com/guimaraes13/NeuralLog}}.
Therefore NeuralLog provides a very flexible approach, it has some limitations, for instance, it is unable of handling propositional predicates in the head of the rules. We also limited the use of facts to predicates with arity up to two, since TensorFlow, in its used version (2.0), cannot handle sparse tensors with more than two dimensions and the use of dense tensors to represent facts with arity bigger than two would easily consume too much memory. On the other hand, predicates of arity bigger than two are useful when used as target predicates, since we can provide more information for each example; and they are also useful in the rules, to combine different information, facilitating the definition of siamese network \citep{jBromley93}, for instance.


We opt to use multiplication to represent the logic \textbf{AND} and sum to represent the logic \textbf{OR} because they have a similar meaning to what we would expect from logic programs and they are already provided by TensorFlow. Furthermore, we allow those functions to be changed in the logic program, given the implementation is provided.

It is important to point out that the first-order logic expressiveness cannot be exactly represented by means of differentiable function operations. In this work, we present a logic language to abstract the structure of neural networks and we try to make the semantics of the neural network as close as possible as the one of the logic program. However, may not be exactly the same.

\section{Experiments} \label{sec:experiments}

In order to demonstrate the flexibility of NeuralLog, we compared it with two distinct relational systems that uses first-order logic to create relational neural networks. In these experiments, we would like to answer two main research questions, in order to address the flexibility of NeuralLog: (\textbf{Q1}) does NeuralLog perform well on \emph{link prediction} tasks? And (\textbf{Q2}) does NeuralLog perform well on \emph{classification} tasks?

In order to answer each question, we compared NeuralLog with other similar systems proposed specifically for each type of task.
The first one is TensorLog \citep{wCohen17}, which uses belief propagation to compile logic theories into neural networks. The second one is RelNN \citep{sKazemi18s}, which learns relational neural networks that use the number of instantiation of logic rules as features to predict the value of a predicate for a given entity.

TensorLog and RelNN are two distinct systems that use logic to build neural networks. On one hand, TensorLog focus at \emph{link prediction}, by retrieving queries of the form $q(a, X)$, where the goal is to find all the entities that are related (as second term) to the entity $a$ (as first term) through the relation $q$. On the other hand, RelNN focus at \emph{classification}, by predicting the value of relations of the form $q(X)$, for a given entity $a$, based on the relations of $a$, defined by logic rules.

First, we compared NeuralLog with TensorLog. Since TensorLog predicts the values of the entities related to a given entity, we ran link prediction experiments in three datasets: the Cora dataset \citep{hPoon07}, and the UWCSE dataset \citep{mRichardson06}, two very popular datasets among ILP systems; and the WordNet dataset \citep{gMiller95}, a frequently used benchmark dataset.

The link prediction task is the task of predicting the entities that relate to a given entity, through a target predicate. To perform this task, we have a set of examples in the form $q(a, x)$, where we give the first entity to the system $a$, and wants to predict the last entity $x$.

Finally, we compared NeuralLog with RelNN in classification tasks in two different datasets: the Yelp dataset 
and the PAKDD15 dataset, both datasets were available with RelNN at \citep{sKazemi18s}. For the Yelp dataset, the task is to predict whether a restaurant serves mexican food through the relation $mexican(X)$; while in the PAKDD15, the task is to predict the gender of a person, based on the movies the person watches, through the relation $male(X)$.

\subsection{Methodology}

\begin{table}
	\center
	\caption{Size of the datasets}
	\begin{tabular}{l|r|rr|rr}
		\toprule
		\multirow{2}{*}{Dataset} 	& \multicolumn{1}{c|}{\multirow{2}{*}{Facts}}	& \multicolumn{2}{c|}{Train}	& \multicolumn{2}{c}{Test}	\\
									&												& \multicolumn{1}{c}{Pos}		& \multicolumn{1}{c|}{Neg} & \multicolumn{1}{c}{Pos} & \multicolumn{1}{c}{Neg} \\
		\midrule
		Cora	& 44,711	& 28,180	& 24,104	& 7,318	& 3,660	\\
		WordNet	& 106,088	& 35,354	& 53,732	& 5,000	& 9,098	\\
		UWCSE	& 3,146		& 90		& 49,622	& 23	& 3,320	\\
		\midrule
		Yelp	& 45,541	& 2,160		& 1,368		& 540	& 342	\\
		PAKDD15	& 127,420	& 5,301		& 18,702	& 1,325	& 4,676	\\
		\bottomrule
	\end{tabular}
	\label{tab:datasets}
\end{table}

We use the same logic theory available with the compared systems for the respective datasets. In our system, for Cora, WordNet and UWCSE, we applied a tanh function to the output of the rules, and also added a bias to the target predicate; then, applying another tanh to the weighted sum of the rules and the bias for the target predicate. Those functions act as activation and output function, respectively.
The theories for the Yelp and PAKDD15 already had activation functions and biases, the only difference is that RelNN used the sigmoid function, while we use the tanh function.

In the case of the UWCSE, we have used a manually written theory provided by Alchemy\footnote{\url{http://alchemy.cs.washington.edu/}}. Since Alchemy support a more complex logic set, we removed rules that uses logic features not support by TensorLog and NeuralLog. The UWCSE dataset also contains facts of arity three, which are not supported by either the systems, we converted them by concatenating two terms that always appears together in the theory. We also added two additional predicates to extract either term, given the concatenated terms.

A bias rule has the form $target(X, Y) \leftarrow b.$, where $b$ is a fact to be learned. Since the variables of the head do not depend on its body, the rule is always true. Then, the rule is summed to other rules with the same head, acting as a bias.

We use adagrad to optimize the mean square error, running for $10$ epochs for each dataset.
We ran each experiment $10$ times and reported the average area under the ROC curve for each dataset. We set the recursion depth to $1$, meaning that the recursive rule is applied only once.
For TensorLog and RelNN, we used the parameters reported at their papers, \citep{wCohen17} and \citep{sKazemi18s}, respectively.
We use holdout with Cora and WordNet; and 5-folds cross-validation with UWCSE, Yelp and PAKDD15, since the datasets were divided that way.

In the WordNet dataset, we used $25\%$ of the facts in the train set and the remaining as KB.
Since the WordNet has no negative examples, we artificially generated approximately $2$ negative examples for each entity appearing in the first position of a target relation. We did this by replacing the second entity of the example by another entity appearing in same relation, which does not appear as positive example, following the Local Closed World Assumption \citep{lGalarraga13}.

The size of the datasets can be seen in Table \ref{tab:datasets}. The UWCSE, Yelp and PAKDD15 lines show the average size of the cross-validation folds.

\subsection{Results}

\begin{table*}
	\center
	\caption{Area under ROC curve for Cora dataset}
	\begin{tabular}{l||c|c||c|c}
		\toprule
		\multirow{2}{*}{Relation}	&	\multicolumn{2}{c||}{Individual}	&	\multicolumn{2}{c}{Together}	\\
		&	NeuralLog	& TensorLog	& NeuralLog	& TensorLog	\\
		\midrule
		S. Author	& $\mathbf{1.0000 \pm 0.0000}$	& $0.9276 \pm 0.0000$	& $\mathbf{0.9273 \pm 0.0867}$	& $0.8853 \pm 0.0019$ \\
		S. Bib	& $0.9174 \pm 0.0073$			& $\mathbf{0.9339 \pm 0.0000}$	& $\mathbf{0.9299 \pm 0.0085}$	& $0.8112 \pm 0.0487$	\\
		S. Title	& $\mathbf{0.8710 \pm 0.0003}$	& $0.5000 \pm 0.0000$	& $\mathbf{0.8728 \pm 0.0046}$	& $0.8442 \pm 0.0274$	\\
		S. Venue	& $\mathbf{0.7810 \pm 0.0077}$	& $0.5000 \pm 0.0000$	& $\mathbf{0.7561 \pm 0.0184}$	& $0.7000 \pm 0.0000$	\\
		\bottomrule
	\end{tabular}
	\label{tab:cora}
\end{table*}

We applied both NeuralLog and TensorLog to each relation of the Cora dataset. When learning each relation individually, NeuralLog achieved better results than TensorLog for the \emph{Same Author} relation, while it is slightly worst for the \emph{Same Bib} relation. Strangely, TensorLog was not able to learn the \emph{Same Title} and \emph{Same Venue} relations in any of the $10$ runs.
We repeated the Cora experiments learning all the relations together. In this setup, NeuralLog outperformed TensorLog in all relations. We reported these results in Table \ref{tab:cora}. It is worth pointing out that TensorLog was able to learn only in $6$ and $1$ out of $10$ runs, for the \emph{Same Title} and \emph{Same Venue} relations, respectively. We reported the average of the successful runs.

In the WordNet dataset, NeuralLog achieved slightly worst results than TensorLog. NeuralLog achieved $\mathbf{0.6593}$ and $\mathbf{0.6679}$ for the weighted and arithmetic average, respectively; of the area under ROC curve of the $18$ relations. Against $\mathbf{0.6604}$ and $\mathbf{0.6766}$ from TensorLog.

In the UWCSE dataset, NeuralLog achieved $\mathbf{0.9509}$ for the area under the ROC curve, while TensorLog was only able to achieve $\mathbf{0.7107}$. It is important to note that we were not able to run TensorLog with the same theory as NeuralLog, since it raised an error when rules containing free variables or rules whose the output variable had no inputs connected to it, for more details, see Appendix \ref{app:uwcse}. Using the same limited theory as TensorLog, NeuralLog achieved $\mathbf{0.7111}$ on area under the ROC curve.

Finally, in comparisons with RelNN, NeuralLog achieved significantly better results in both Yelp and PAKDD15 datasets. NeuralLog achieved $\mathbf{0.7652}$ and $\mathbf{0.8052}$, against $\mathbf{0.7454}$ and $\mathbf{0.7562}$ from RelNN, for the Yelp and PAKDD15, respectively.

In most cases, NeuralLog achieved better results than the other systems, in tasks they were designed for, even using the same logic theory. This might be due to better optimisation algorithm, and, possibly, better initial parameters.
Nevertheless, our experiments show that NeuralLog is flexible enough to successfully represent other systems based on logic theories and neural networks. Thus, we can affirmatively answer both questions \textbf{Q1} and \textbf{Q2}.

\section{Conclusion} \label{sec:conclusion}

We present NeuralLog, a first-order logic language that is compiled into a neural network. Experiments show that NeuralLog can represent both TensorLog \citep{wCohen17} and RelNN \citep{sKazemi18s} programs, achieving even better results in the Cora, UWCSE, Yelp and PAKDD15 datasets; and comparable results in the WordNet dataset. NeuralLog is also capable of representing MLPs and other more complex neural network structures (such as cosine similarity) due to its capability of handling numeric attributes and functions; as well as rules with free variables, which improved the results in the UWCSE dataset. 

The main goal of NeuralLog is to bridge advances in both deep learning and logic programming, two distinct fields of machine learning that can complement each other.
As future work, we would like to explore ILP algorithms in order to learn the logic programs (and, consequently, the structure of the neural networks) from examples.


\subsubsection*{Acknowledgment}

Victor Guimarães was financed by the Portuguese funding agency, FCT - Fundação para a Ciência e a Tecnologia through Ph.D. scholarships 2020.05718.BD.

\bibliography{references}
\bibliographystyle{iclr2021_conference}


\appendix
\section{Rule Paths Algorithms and Examples} \label{sec:rule_algorithms}

\begin{algorithm*}
    \caption{Find clause paths: Algorithm to find the paths between the sources and the destination terms of a clause and the disconnected literals} \label{alg:find_clause_paths}
    \begin{algorithmic}[1]
    \Statex
    \Require{The clause ($clause$); and the destination term index ($dest\_index$);}
    \Ensure{The paths from the source terms to destination; and the disconnected ground literals;}
    \Statex
	\Function{find\_clause\_paths}{$clause, dest\_index$} 
		\State $sources \gets clause.head.terms$
		\State $compute\_reverse \gets True$
		\State $destination \gets source[dest\_index]$
		\If{$|sources| > 1$}
			\State $sources.remove[dest\_index]$
		\Else
			\State $compute\_reverse \gets False$
		\EndIf
		\State $all\_paths \gets \{\}$
		\State $all\_visited\_nodes \gets \{\}$
		\State $sources\_set \gets$ set($sources$)
		\Statex

		\For{\textbf{each} $source \in sources$}
			\State $visited\_nodes \gets \{\}$
			\State $paths, visited\_nodes \gets$ find\_paths($clause, source, destination, visited\_nodes, sources\_set$)
			\State $all\_paths \gets all\_paths \cup paths$
			\State $all\_visited\_nodes \gets all\_visited\_nodes \cup visited\_nodes$
		\EndFor
		\Statex

		\If{$compute\_reverse$ \textbf{and} $clause.body \nsubseteq all\_visited\_nodes$}
			\State $destination \gets source$
			\For{\textbf{each} $source \in sources$}
				\State $sources\_set \gets$ set($clause.head.terms$)$ - \{ destination \}$
				\State $backwards\_paths, all\_visited\_nodes \gets$ \par
				\hfill find\_paths($clause, source, destination, all\_visited\_nodes, sources\_set$)
				\For{each $backwards\_path \in backwards\_paths$}
					\State $path \gets $ reverse($backwards\_path$)
					\State $all\_paths \gets all\_paths \cup \{ path \}$
				\EndFor
			\EndFor
		\EndIf
		\Statex

		\State $disconnected\_literals \gets$ get\_disconnected\_literals($clause, all\_visited\_nodes$) \par
		\hfill \Comment{gets the ground literals that does not bellong to any path}
		\\ \Return $all\_paths, disconnected\_literals$
	\EndFunction
  \end{algorithmic}
\end{algorithm*}

\begin{algorithm*}
    \caption{Find paths: algorithm to find the paths between the source and destination terms} \label{alg:find_paths}
    \begin{algorithmic}[1]
    \Statex
    \Require{The clause ($clause$); the source term ($source$); the destination term ($destination$); the visited nodes ($visited\_nodes$); and all source terms ($sources\_set$);}
    \Ensure{The paths from the source terms to destination; and the visited nodes}
	\Statex
	\Function{find\_paths}{$clause, source, destination, visited\_nodes, sources\_set$} 
		\State $completed\_paths \gets []$
		\State $partial\_paths \gets [[source]]$
		\State $edge\_literals \gets$ get\_non\_loop\_literals($clause$)
		\While{$|partial\_paths| > 0$}
			\State $size \gets |partial\_paths|$
			\For{$i \gets 0; i < size; i++$}
				\State $path \gets$ pop\_left($partial\_paths$)
				\If{end($path$) $= destination$}
					\State $completed\_paths \gets completed\_paths + [path]$
				\State \textbf{continue to next for iteration}
				\EndIf
				\State $not\_added\_path \gets True$
				\For{\textbf{each} $literal \in$  get\_literal\_with\_term($edge\_literals$, end($path$))}
					\State $new\_end \gets$ compute\_end\_term($path, literal$)
					\If{$new\_end \in path$ \textbf{or} $new\_end \in sources\_set$} 
						\State \textbf{continue to next for iteration} \Comment{path comes back to itself or to another input}
					\EndIf
					\State $new\_path \gets path + [literal, new\_end]$
					\If{$new\_end = destination$}
						\State $completed\_paths \gets completed\_paths + [new\_path]$
					\Else
						\State $partial\_paths \gets partial\_paths + [new\_path]$
					\EndIf
					\State $visited\_nodes \gets visited\_nodes \cup {literal}$
					\State $not\_added\_path \gets False$
				\EndFor
				\If{$not\_added\_path$}
					\State $completed\_paths \gets completed\_paths + [path + [ANY, destination]]$
				\EndIf
			\EndFor
		\EndWhile
		\State $completed\_paths \gets$ append\_loops\_to\_paths($clause, completed\_paths$)
		\\ \Return $completed\_paths, visited\_nodes$
	\EndFunction
  \end{algorithmic}
\end{algorithm*}

\subsection{Rule Examples}

The following examples show two examples of the expressiveness of NeuralLog, each of which with a single rule: (1) a rule where the prediction is weighted by an attribute of the entity; and (2) a rule to aggregate a value, based on the relations of an entity.

\begin{example}
	consider the following rule:
	\begin{equation*}
		influence(X, Y) \leftarrow friends(X, Y) \wedge age(X, A) \wedge weight.
	\end{equation*}
	There will be two possible paths between $X$ and $Y$: (1) $X \rightarrow friends \rightarrow Y$; and (2) $X \rightarrow age \rightarrow A \rightarrow any \rightarrow Y$.

	From the first path, we would get a resulting vector $Y'$ with all non-zero values representing the entities that are friends of $X$. From the second path, we would have a vector $Y''$ with the age of $X$ in every position. The final vector, $Y$ would then be given by the element-wise multiplication $Y' \odot Y''$ and then by the scalar $weight$.

	Thus, this rule means that a person $X$ influences their friends $Y$ proportionally to his/her. Additionally, we can learn the weight $weight$ for the rule, independent of any entity. Similar kind of rules could be used to retrieve numeric attributes from entities and combine them to create a \emph{Multilayer Perceptron} (MLP).
\end{example}

\begin{example}
	consider the following rule:
	\begin{multline*}
		mean\_age\_of\_friends(X) \leftarrow \\ friends(X, Y) \wedge age(Y, A) \wedge mean(A).
	\end{multline*}
	This rule have a single path, starting from $X$: $X \rightarrow friends \rightarrow Y \rightarrow age \rightarrow A \rightarrow mean \rightarrow A$.

	This path starts by computing vector $Y$, which represents the friends of $X$ by multiplying $X$ by $friends$. Then, it obtains the vector $A$, containing the ages of the friends of $Y$, by multiplying $Y$ by $age$. Finally, it applies the function $mean$ to the vector $A$, which computes the mean of all non-zero entries for $A$, resulting in the average of the age of the friends of $X$.
\end{example}

\subsection{Cosine Example}

Table \ref{tab:cos_example} shows an example of a program to learn latent features to the entities in the knowledge base and compute the similarity of those entities, based on the cosine of the angle between their latent feature vectors.

\begin{table*}
	\center
	\caption{Cosine similarity example} \label{tab:cos_example}
	\begin{tabular}{ll}
		\toprule
		$h1(X, Y) \leftarrow l1(X) \wedge l1(Y).$		& $square\_sum(X) \leftarrow l1(X) \wedge l1(X).$	\\
		$\vdots$										& $\vdots$	\\
		$h10(X, Y) \leftarrow l10(X) \wedge l10(Y).$ 	& $square\_sum(X) \leftarrow l10(X) \wedge l10(X).$	\\
														& $norm(X) \leftarrow square\_sum(X) \wedge square\_root(X).$\\
		$num(X, Y) \leftarrow h1(X, Y).$				& $den(X, Y) \leftarrow norm(X) \wedge norm(Y).$\\
		$\vdots$										& $inv\_den(X, Y) \leftarrow inv\_den(X, Y) \wedge inverse(Y).$	\\
		$num(X, Y) \leftarrow h10(X, Y).$				& $similarity(X, Y) \leftarrow num(X, Y) \wedge inv\_den(X, Y).$	\\
		\bottomrule
	\end{tabular}
\end{table*}

In this program: $square\_root$ and $inverse$ are numeric functions to compute the element-wise square root and inverse value of the elements in a vector, respectively; the predicates $l1$ to $l10$ are predicates to be learned; and $similarity$ is the target predicate that computes the cosine similarity between the vectors of $X$ and $Y$, where the components of the vectors are stored in the predicates $l1$ to $l10$, for each entity in the KB. The cosine similarity between two vectors is shown in Equation \ref{eq:cos_sim}. From a set of examples of similar entities, we can train the network to learn the weights of the predicates $l1$ to $l10$, representing $10$ latent features for each entity.

\begin{equation} \label{eq:cos_sim}
	similarity(\mathbf{x}, \mathbf{y}) = \frac{<\mathbf{x}, \mathbf{y}>}{\norm{\mathbf{x}}\norm{\mathbf{y}}}
\end{equation}

The rules $h_i(X, Y) \leftarrow \dots$ calculates, for a given $X$, a vector with the value of the latent feature $i$ of $X$ times the latent feature $i$ of $Y$, for each possible entity $Y$. This is due the syntax of the $any/n$ predicate.
The rules $num(X, Y) \leftarrow \dots$ sums all the results from the rules $h_i(X, Y) \leftarrow \dots$, computing the numerator part of the cosine similarity. In a similar way, the rule $den(X, Y) \leftarrow \dots$ calculates the denominator, by multiplying the norm of the latent features of the input entity $X$ for each possible output entity $Y$. Finally, $similarity(X, Y) \leftarrow \dots$ computes the cosine similarity, by multiplying the numerator by the inverse of the denominator, given by the rule $inv\_den(X, Y) \leftarrow inv\_den(X, Y) \wedge inverse(Y)$.

\section{UWCSE Theory} \label{app:uwcse}

We also experimented with the UWCSE dataset \citep{mRichardson06}, a well-known dataset in ILP. We applied a simplified version of the theory provided by Alchemy \footnote{\url{http://alchemy.cs.washington.edu/}}. Since the original UWCSE dataset contains two predicates of arity $3$, we have to convert them to binary predicates. The original predicates were $taughtby(course, person, season)$ and $ta(course, person, season)$. We transformed these predicates to binary by creating a single constant $k$, by concatenating the $course$ and $season$, since they appears together in the theory. We also added a new predicate $hascourse(k, course)$, to extract the course constant from the concatenated keys $k$. 

Table \ref{tab:uwcse-theory} shows the theory used in the experiments with the UWCSE dataset. The underlined part represents the subset of the theory used by TensorLog, since it was unable to deal with rules containing free variables (variables that appears only once in the rule), either in the head or in the body. It also does not accept rules whose output variable has no inputs connected to it, for example $advisedby(X1, X2) \leftarrow student(X1) \wedge professor(X2)$. Using the full theory, NeuralLog achieved $0.9509$ for the area under the ROC curve. Since TensorLog has to use a simpler theory, it was only able to achieve $0.7107$ on the area under ROC curve. Using the same limited theory as TensorLog, NeuralLog achieved $0.7111$ on the same metric.

\begin{table*}
	\center
	\caption{UWCSE Theory} \label{tab:uwcse-theory}
	\begin{tabular}{l}
		\toprule
		$\underline{advisedby(X1, X2) \leftarrow taughtby2(K, X2) \wedge ta2(K, X1)} \wedge hascourse(K, X3)$ \\ 
		\hfill $\wedge \ courselevel(X3, X5) \wedge tanh(X2) \wedge w(w1).$ \\
		\\
		$advisedby(X1, X2) \leftarrow taughtby2(K, X2) \wedge ta2(K, X1)$ \\
		\hfill $\wedge inphase(X1, X5) \wedge tanh(X2) \wedge w(w2).$ \\
		\\
		$\underline{advisedby(X1, X2) \leftarrow publication(X3, X1) \wedge publication(X3, X2) \wedge student(X1)}$ \\
		\hfill $\underline{\wedge \ professor(X2)} \wedge tanh(X2) \wedge w(w3).$ \\
		\\
		$\underline{advisedby(X1, X2) \leftarrow publication(X3, X1) \wedge publication(X3, X2)} \wedge tanh(X2) \wedge w(w4).$ \\
		\\
		$\underline{advisedby(X1, X2) \leftarrow publication(X3, X1) \wedge publication(X3, X2) \wedge student(X1)}$ \\
		\hfill $\wedge \ tanh(X2) \wedge w(w5).$ \\
		\\
		$advisedby(X1, X2) \leftarrow student(X1) \wedge tanh(X2) \wedge w(w6).$ \\
		$advisedby(X1, X2) \leftarrow professor(X2) \wedge hasposition(X2, X3) \wedge tanh(X2) \wedge w(w7).$ \\
		$advisedby(X1, X2) \leftarrow professor(X2) \wedge tanh(X2) \wedge w(w8).$ \\
		$advisedby(X1, X2) \leftarrow b(advisedby).$ \\
	\bottomrule
	\end{tabular}
\end{table*}

\end{document}